\documentclass[sigconf]{acmart}

\usepackage{booktabs} 

\setcopyright{acmcopyright}

\acmDOI{https://doi.org/10.48550/arXiv.2312.01661}

\acmISBN{979-8-4007-0243-3/24/04}

\acmConference[SAC'24]{ACM SAC Conference}{April 8 –April 12, 2024}{Avila, Spain}
\acmYear{2024}
\copyrightyear{2024}

\acmArticle{4}
\acmPrice{15.00}

\usepackage{cleveref}
\usepackage{color}
\usepackage{colortbl}
\usepackage{graphicx}
\usepackage{makecell}
\usepackage[ruled,linesnumbered]{algorithm2e}
\usepackage{multirow}
\usepackage{soul}
\usepackage{xcolor}

\newcommand{\dataset}{\textsc{PRE-UMATH}}
\newcommand{\topicmath}{\textsc{TopicMath}}
\usepackage{listings}

\usepackage{tcolorbox}

\begin{document}
\title{ChatGPT as a Math Questioner? Evaluating ChatGPT on Generating Pre-university Math Questions} 
  
\renewcommand{\shorttitle}{SIG Proceedings Paper in LaTeX Format}

\author{Phuoc Pham Van Long}
\authornote{Equal contribution.}
\affiliation{%
  \institution{Nanyang Technological University}
  \country{Singapore}}
\email{phuoc002@e.ntu.edu.sg}

\author{Duc Anh Vu\footnotemark[1]}
\affiliation{%
  \institution{Nanyang Technological University}
  \country{Singapore}}
\email{ducanh001@e.ntu.edu.sg}

\author{Nhat M. Hoang\footnotemark[1]}
\affiliation{%
  \institution{Nanyang Technological University}
  \country{Singapore}}
\email{nhat005@e.ntu.edu.sg}

\author{Xuan Long Do\footnotemark[1]}
\affiliation{%
  \institution{National University of Singapore}
  \country{Singapore}}
\email{xuanlong.do@comp.nus.edu.sg}

\author{Anh Tuan Luu}
\authornote{Corresponding author.}
\affiliation{%
  \institution{Nanyang Technological University}
  \country{Singapore}}
\email{anhtuan.luu@ntu.edu.sg}

\renewcommand{\shortauthors}{Pham et al.}
\newcommand{\nhat}[1]{\textcolor{red}{(Nhat: \textcolor{blue}{#1})}}
\newcommand{\vda}[1]{\textcolor{purple}{(VDA: \textcolor{blue}{#1})}}
\newcommand{\llong}[1]{\textcolor{orange}{(Long: #1)}}
\newcommand{\phuoc}[1]{\textcolor{red}{(PHUOCDZ: \textcolor{blue}{#1})}}

\begin{abstract}
Mathematical questioning is crucial for assessing students' problem-solving skills. Since manually creating such questions requires substantial effort, automatic methods have been explored. Existing state-of-the-art models rely on fine-tuning strategies and struggle to generate questions that heavily involve multiple steps of logical and arithmetic reasoning. Meanwhile, large language models (LLMs) such as ChatGPT have excelled in many NLP tasks involving logical and arithmetic reasoning. Nonetheless, their applications in generating educational questions are underutilized, especially in the field of mathematics. To bridge this gap, we take the first step to conduct an in-depth analysis of ChatGPT in generating pre-university math questions. Our analysis is categorized into two main settings: \emph{context-aware} and \emph{context-unaware}. In the context-aware setting, we evaluate ChatGPT on existing math question-answering benchmarks covering elementary, secondary, and ternary classes.  In the context-unaware setting, we evaluate ChatGPT in generating math questions for each lesson from pre-university math curriculums that we crawl. Our crawling results in {\topicmath}\footnote{Our codes and data are publicly available at \url{https://github.com/dxlong2000/ChatGPT-as-a-Math-Questioner}.}, a comprehensive and novel collection of pre-university math curriculums collected from $121$ math topics and $428$ lessons from elementary, secondary, and tertiary classes. Through this analysis, we aim to provide insight into the potential of ChatGPT as a math questioner\footnotemark[1].

\end{abstract}

%
%



\maketitle

\section{Introduction}
Math problems are essential educational tools for evaluating students' logical and problem-solving abilities \cite{verschaffel2020word, hendrycks2021measuring}. Engaging students in answering those expert-designed questions has been shown to improve their learning outcomes \cite{prince2004does,karpicke2008critical}. Nonetheless, manually crafting such questions demands substantial human effort and expertise, making it time-consuming, non-generalizable, and impractical for scalability \cite{lu2021expert}. Therefore, automatic tools to generate mathematical questions have received growing attention \cite{liu2021mathematical,wang-etal-2021-math}. Existing state-of-the-art frameworks primarily rely on fine-tuning strategies \cite{10.5220/0004795300140025,wang-etal-2021-math,shridhar2022automatic,9723526}. However, these approaches are criticized for their limitations in generating questions that necessitate multi-step reasoning \cite{kulshreshtha2022reasoning}. Recent progress in large language models (LLMs), like ChatGPT \cite{openai2022chatgpt}, has garnered significant interest and demonstrated remarkable efficacy in numerous natural language processing (NLP) tasks through the use of prompts. Nevertheless, their potential and benefits in crafting educational questions, especially within mathematics, remain underinvestigated.

In this work, we take the first step to conduct an in-depth analysis of the potential of applying ChatGPT in automatically generating pre-university math questions. We categorize our analysis into two main scenarios: \emph{(1) context-aware}, where the model is given a context to generate math questions either with or without an expected answer, and \emph{(2) context-unaware}, where the model generates math questions based solely on an instructional prompt. Under the context-aware setting, ChatGPT is evaluated on 3 math question-answering benchmarks from elementary, secondary, and ternary classes respectively. In context-unaware scenarios, where no prior context is available, assessing ChatGPT is more challenging due to significant variations in model performance based on different instructional prompts. Nonetheless, this setting is more realistic and helpful since teachers may not have any contexts or stories beforehand to ask for generating math questions.  

In addition, our evaluation reveals that the performance of the model varies when generating questions in different math topics. Therefore, to exhaustively evaluate the model in the context-unaware setting, we hire expert students who are high-school national math olympians from universities, to crawl $428$ math lessons from Khan Academy\footnote{https://www.khanacademy.org/} with their mathematical definitions and exemplary problems from $121$ math topics covering most of the topics from grade $1$-st to tertiary classes. We then instruct ChatGPT to generate question-answer pairs for each lesson, given desired difficulty levels. Through our analysis, we derive a number of worthy findings. Our contributions are summarized below:

\emph{(i)} We are the first to conduct a comprehensive analysis of the feasibility of leveraging ChatGPT in generating pre-university math questions. 

\emph{(ii)} We study two main settings in generating mathematical questions. We further dive our evaluation deeply into a large number of math topics and lessons covering most from pre-university classes.

\emph{(iii)} We contribute \topicmath{}, a novel and comprehensive collection of expert-authored pre-university math curriculums. 

\emph{(iv)} We provide $11$ findings about the capability of ChatGPT in generating pre-university math questions. We hope these findings can offer good insights for teachers \& researchers in utilizing modern AI technologies like ChatGPT for serving educational purposes.

\section{Related Work}
\subsection{Large Language Models \& Prompting}
Recently, LLMs have shown remarkable zero-shot and few-shot abilities in various language generation contexts \cite{brown2020language, wei2022finetuned, ouyang2022training}. However, they still face challenges in more complex tasks like mathematical reasoning \cite{gao2022pal, imani2023mathprompter}, often requiring expensive computational resources for fine-tuning. To address this, researchers have been exploring novel prompting methods to instruct LLMs in these tasks, including chain-of-thought (CoT) prompting \cite{weichain}. This enables LLMs to perform intermediate reasoning steps, significantly enhances LLMs' reasoning abilities, especially for complex mathematical and decision-making tasks.

\subsection{Pre-university Math Problems Generation}\label{ssec:related-math-problem}
Pre-university math problems have received increasing attention from the AI research community, with benchmarks such as SVAMP \cite{svamp} for elementary-level math, secondary school-level GSM8K \cite{cobbe2021training} offers diverse solution templates, and the MATH \cite{hendrycks2021measuring} dataset provides complex reasoning for tertiary/olympiad problems along with step-by-step solutions. Recently, interest has grown in other tertiary math topics like geometry problems and mathematical theorem proving \cite{sachan-etal-2019-discourse, chen-etal-2022-unigeo}. Additionally, automatic question generation (QG) in education has gained attention for enhancing teaching activities \cite{Kurdi2019ASR}. Additionally, in education, QG has gained attention with the use of LLMs, particularly ChatGPT, has gained significant interest for generating practice questions in various subjects \cite{Wang2022TowardsHE, Kasneci2023ChatGPTFG}. However, its potential for generating pre-university mathematics problems remains largely unexplored. This study, therefore, evaluates ChatGPT's performance using three well-established datasets: SVAMP, GSM8K, and MATH, covering pre-university grades and various difficulty levels.

\section{Problem Formulation}
We study  ChatGPT\footnote{https://openai.com/blog/chatgpt} on generating math problems in both \textbf{context-aware} and \textbf{context-unaware} settings across various pre-university difficulty levels, including elementary, secondary, and tertiary.

\paragraph{$\bullet$ Context-aware} We evaluate models in both \emph{answer-aware} and \emph{answer-unaware} modes. In the \emph{answer-aware} setting, we provide context $C$ and evaluate the models by generating math questions, with each sample represented as $(C, Q, A)$, where $C$ is the context, $Q$ is the question, and $A$ is the answer. The models are then fine-tuned/run inference to generate $Q$ given $C$ and $A$. In the \emph{answer-unaware} setting, models generate questions conditioned solely on $C$, with $A$ being unavailable.

\paragraph{$\bullet$ Context-unaware} The absence of context $C$ poses a unique challenge for assessing ChatGPT's math problem generation capabilities. Nonetheless, this scenario is essential, as teachers often seek to prompt language models like ChatGPT for math questions without prior context. To address this, we manually collect math curricula for three pre-university levels and propose a prompting framework to create \dataset, a novel dataset with $16K$ question-answer pairs spanning $121$ pre-university math topics and $428$ lessons. Our evaluation of \dataset\ provides valuable insights into ChatGPT's math question generation capability.

\section{Context-aware Methodology}
\subsection{Fine-tuning Baselines} We fine-tune the baselines to generate the question $Q$, given the context $C$ with or without the expected answer $A$ by concatenating the input in the format: \texttt{Context: C} [with/without] \texttt{Answer: A}. The model then learns to generate $Q$.

\subsection{Prompting ChatGPT}  
We prompt ChatGPT to generate a math question using $C$ with or without $A$. Empirical experiments in Table \ref{constraints_compa} demonstrate that imposing constraints produces questions closer to ground truth. Hence, we propose the following constraints for this task. To ensure coherence and comparison with groundtruth questions, we instruct ChatGPT to generate concise questions \textbf{(1) without excessive context repetition}. This constraint minimizes disparities with the ground-truth question, improving fluency (e.g., before: \texttt{[Context] [Question] [Redundant Context]}; after: \texttt{[Context] [Question]}). To maintain consistency, we emphasize that the generated question should \textbf{(2) match the tense} of the provided context. This constraint helps to ensure that the question appears grammatically correct and coherent within the given context (e.g., before: \texttt{[Past-tense Context] [Present-tense Question]}; after: \texttt{[Past-tense Context] [Past-tense Question]}). In order to promote brevity and clarity in the generated questions, we set a \textbf{(3) word limit} of no more than 20 words.

\begin{table*}[t!]

\centering 
\small
\scalebox{.8}{
\begin{tabular}{lc|cccc|cccc|cccc}
\toprule
\multirow{2}{*}{\textbf{Model}} & \multirow{2}{*}{\textbf{Mode}} & \multicolumn{4}{c}{\textbf{SVAMP}}  & \multicolumn{4}{c}{\textbf{GSM8K}} & \multicolumn{4}{c}{\textbf{MATH}}\\
 &  & B-4 & R-L & Meteor & BERTScore & B-4 & R-L & Meteor & BERTScore & B-4 & R-L & Meteor & BERTScore \\
\midrule
ChatGPT &  zero-shot & 10.13 & 34.52 & 54.04  & 90.93 & 11.75 & 34.87 & \textbf{53.17} & 90.71 & 18.95 & 39.63 & \textbf{48.35} & 88.87\\
\makecell[l]{ChatGPT \\+ CI + TM + WL} &  zero-shot & \textbf{23.57} & \textbf{57.82} & \textbf{65.43} & \textbf{93.97} & \textbf{12.19} & \textbf{38.55} & 44.27 & \textbf{91.49} & \textbf{19.48 }& \textbf{42.12} & 46.36 & \textbf{89.11}\\
\bottomrule
\end{tabular}
}
\caption{\small{Comparisons between performances of ChatGPT with and without constraints: Contextual Independence, Tense Matching and Word Limit}}
\label{constraints_compa}
\normalsize
\vspace{-5mm}
\end{table*}

\section{Context-unaware Methodology} 
\subsection{\topicmath{} Creation}
\label{subsec:curri}
Math problems cover a wide range of math topics with varying levels of logical complexity. Our preliminary experiments suggest that the performance of ChatGPT on different math topics might not be the same. To evaluate its capability thoroughly, we specifically examine the model on multiple topics and lessons that we crawl in elementary, secondary, and tertiary classes. After exhaustive searches, since there are no standard math curriculums consisting of topics and lessons for these classes internationally, we choose Khan Academy\footnote{https://www.khanacademy.org/} as our source of curriculums since it has been well-recognized both in the US and internationally. Our crawling results in {\topicmath}, a comprehensive and
novel collection of pre-university math curriculums collected from
$121$ math topics and $428$ lessons from elementary, secondary, and
tertiary classes. Its statistics are presented in \Cref{data-collecting-khan}.

\begin{table}[ht]{
\centering
\raggedright
\textcolor{white}{-----------------------------------------------------------------------------------------}}
\scriptsize
\begin{tabular}{ccccccc}
\cline{1-6}
{\textbf{Class}} & \makecell{\textbf{\#collected }\\ \textbf{topics}} & \makecell{\textbf{\#total} \\\textbf{topics}} &  \makecell{\textbf{\#collected }\\ \textbf{subtopics}}  & \makecell{\textbf{\#total }\\ \textbf{subtopics}} & \makecell{\textbf{\%removed }\\ \textbf{subtopics}} \\
\cline{1-6}
Grade 1 & 2 & 3 & 4 & 17 & 76.47\%         \\
Grade 2 & 6 & 8 & 24 & 33 &  27.27\%   \\
Grade 3 & 11 & 14 & 36 & 65 & 44.62\%     \\
Grade 4 & 11 & 14 & 20 & 75 & 73.33\%      \\
Grade 5 & 15 & 16 & 30 & 64 & 52.13\%     \\
Grade 6 & 11 & 11 & 30 & 74 & 59.46\%   \\
Grade 7 & 7 & 7 &23 & 47 & 51.06\%   \\
Grade 8 & 7 & 7 &32 & 52 & 38.46\%   \\
 \makecell{High\\ school} & 51 & 81 & 229 & 476 & 51.89\%   \\
\cline{1-6}
Total & 121 & 161 & 428 & 903 &  52.60\% \\ 
\cline{1-6} \\
\end{tabular}
\caption{{\topicmath} statistics: Topics and Subtopics Collected by Grade Level} \label{data-collecting-khan}
\vspace{-7mm}
\end{table}

\paragraph{(1) Curriculum Collection}  We hire six undergraduate students in mathematics who achieved high-school national mathematical olympiad medals. They are divided into three groups, each group consists of two students. Students in each group are instructed as follows to collect the math curriculums from Khan Academy. First, they are instructed to collect all the math topics (chapters' titles) and lessons' titles from $14$ courses in Khan Academy, ranging from elementary school to tertiary. In addition to the titles, students are also asked to collect one exemplary question per lesson in the \textbf{Example} section, rate its difficulty following our definitions in \Cref{appdx:human-rating}, and the lessons' definitions from the  \textbf{About} section or the \textbf{FAQ} and \textbf{Review} sections at the end of each chapter. If students could not find any definition in the above sections, they were asked to attempt to find the lesson's definition from the introductory \textbf{Video}. If the students could not find an appropriate definition or example for a lesson, the lesson would not be collected. Among the 14 courses from Khan Academy available, spanning from elementary school to tertiary, the rate of removed lessons is 52.60\% (\Cref{data-collecting-khan}) 

\begin{table}[]
\label{daposedit}    
\vspace{10pt}
\begin{tabular}{l}
\includegraphics[width=0.35\textwidth]{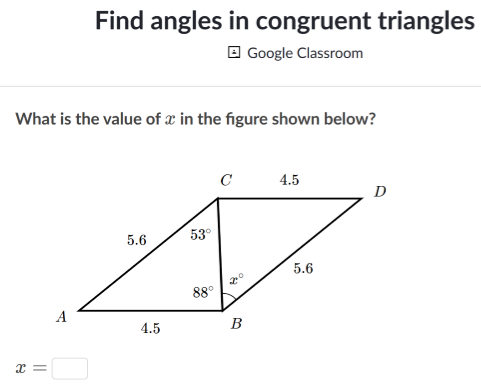}
\end{tabular}

\begin{flushleft}
\textcolor{blue}{Source: Khan Academy}\\
\textcolor{blue}{Question}: What is the value of x in the figure shown below: ... 
\textcolor{blue}{Edited ChatGPT's answer}: Based on the information given.. \textcolor{green}{and Congruence of Triangles - SSS} to ... \textcolor{green}{According to the Congruence of Triangles - SSS ... Hence, angle ABC = angle BCD = 88} Substituting ... + \textcolor{red}{\st{90}} \textcolor{green}{88} degrees + 39 degrees = 180 degrees. Therefore, the value of x in the figure is \textcolor{red}{\st{51}} \textcolor{green}{53} degrees.
\end{flushleft} 

\caption{Data collecting and annotating process with the edit rate of 12.75\%. \textcolor{red}{Red} denotes the deleted part of ChatGPT's answer,  \textcolor{green}{green} denotes the corrected.}
\label{tab:post-edit}
\vspace{-8mm}
\end{table}

\paragraph{(2) Create Examples' Answers}  After getting topics, lessons, definitions, and exemplary questions with their difficulties, we ask annotators to prompt ChatGPT via zero-shot Chain-of-Thought \cite{Kojima2022LargeLM} to obtain the questions' explainable solutions. These solutions are then reviewed and edited as needed. As per the data presented in \Cref{tab:post-edit}, the average edit rate in the token level stands at 12.74\%.

\paragraph{(3) Curriculum Expert Verification}  In our final step, we hire three educators who have degrees in education and currently are math teachers in elementary, secondary, and tertiary schools. They are invited to verify the correctness and appropriateness subjectively of all the collected topics, lessons, definitions, and question-answer pairs with their difficulties. If any collected data is found to be inappropriate or theoretically incorrect, educators have the option to edit or discard it. We found their approval rate of 87.70\%.  Finally, we collect 121 topics and 428 lessons with 428 examples. We name this dataset as \topicmath{}.

\subsection{\topicmath{} Analysis}

\paragraph{Topic \& Subtopic Distribution}
We observe that the number of grade 1 topics collected is the smallest since its difficulty levels are not diverse and the number of mathematical operators and methods is limited, there are fewer collected grade-1-level math questions compared to other levels. Meanwhile, grade 5 has the highest number of topics since a significant number of grade-5-level math questions proved to be highly compatible with our collection criteria and constraints.

\paragraph{Removal Ratio Analysis} In the process of collecting math definitions, questions, and answers in grade 1, our annotators observe the absence of definitions and an overabundance of similar question types. Consequently, a significant portion of grade-1-level questions had to be excluded from our collection due to the stringent criteria and constraints we employ.

\subsection{Prompting ChatGPT to Generate Educational Questions from Math Topics}
\vspace{-3mm}

\begin{algorithm}
\SetKwProg{try}{try}{:}{}
\SetKwProg{except}{except}{:}{end}
\small
\DontPrintSemicolon
\caption{Prompting ChatGPT to generate math questions} \label{alg:answer-unaware-qa-generation}
\textbf{Input: } $course$, $topic$, $subtopic$, $gen\_limit$\\
\textbf{Initialize: } \\

$q\_pool$ = set() \\
\For{$lesson$ in $course[topic][subtopic]$}{
    $definition$ = $lesson["definition"]$ \\
    $qa\_queue$ = $lesson["qa\_queue"]$ \\
    \For{$i$ in range($gen\_limit$)}{
        random.shuffle($qa\_queue$) \\ 
        $qa\_pair$ = $qa\_queue[0]$ \\
        $difficulty$ = $qa\_pair["difficulty"]$ \\
        $prompt$ = create\_prompt($topic$, $subtopic$, $definition$, $difficulty$, $qa\_pair$) \\ 
        $q$, $a$ = get\_chatgpt\_answer($prompt$) \\
        \If{not filter\_question($q$, $q\_pool$)}{
            $qa\_queue$.push(($q$, $a$)) \\
            $q\_pool$.push($q$)
        }
    }
}

\end{algorithm}
\vspace{-5mm}

\begin{table}[h]
\scriptsize
\begin{tabular}{cccccccccc}

\cline{1-9}

\multirow{2}{*}{\textbf{Class}} & \multirow{2}{*}{\textbf{\#topics}} & \multirow{2}{*}{\textbf{\#subtopics}} &  \multirow{2}{*}{{\makecell{\textbf{\#QA}\\ \textbf{pairs}}}} & 
\multicolumn{5}{c}{\textbf{Difficulty distribution}}\\ 
\multicolumn{4}{c}{} & \textbf{1} & \textbf{2} & \textbf{3} & \textbf{4} & \textbf{5}\\
\cline{1-9}
Grade 1 & 2 & 4 & 61 & 61 & 0 & 0 & 0 & 0        \\
Grade 2 & 6 & 24 & 327 &  327  & 0 & 0 & 0 & 0   \\
Grade 3 & 11 & 36 & 506 & 506 & 0 & 0 & 0 & 0     \\
Grade 4 & 11 & 20 & 350 & 0 & 350   & 0 & 0 & 0  \\
Grade 5 & 15 & 30 & 558 & 0 & 558  & 0 & 0 & 0    \\
Grade 6 & 11 & 30 & 1165 & 0 & 0 & 1165 & 0 & 0 \\
Grade 7 & 7 & 23 & 853 & 0 & 0 & 853 & 0 & 0\\
Grade 8 & 7 & 32 & 1231 & 0 & 0 & 1231 & 0 & 0\\
 \makecell{High\\ school} & 51 & 229 & 11032 & 0 & 0 &  60 & 8796& 2176\\
\cline{1-9}
Total & 121 & 428 & 16083 &  894 & 908 & 3309 & 8796 & 2176 \\ 
\cline{1-9} \\

\end{tabular}
\caption{{\dataset} statistics by grades.} \label{tab:diffdist}
\label{table2}
\vspace{-10mm}
\end{table}

We prompt ChatGPT to generate QA pairs from \topicmath{} for our evaluation purposes. Our inference strategy involves using prompts that promote diversification in tokens, topic alignment, and difficulty. The algorithm is presented in Algorithm~\ref{alg:answer-unaware-qa-generation}. Specifically, we create a list of generated QA pairs for each lesson in \topicmath. Given a lesson, our prompt consists of its definition and the topic's name it belongs to, and one demonstration randomly selected from its list of generated QA pairs. After getting the newly generated QA pair, we accept it if its question has a ROUGE-L score less than $0.7$ with any of all the questions generated from all the lessons, otherwise, we filter it out. To promote token diversity in generating educational questions, we utilize two strategies.  For grades 1-8, we ask ChatGPT to enrich the generated questions by providing objects and stories via adding \texttt{"You could introduce characters, objects
or scenarios to make your math problem context more diverse in terms of token"} to the prompts. For tertiary classes, the problems might be complicated and require more rigorous and abstract thinking. Therefore, instead of requiring a real-life context, we instruct the model to introduce more variables in naming the objects via supplementing \texttt{"Your questions are required to be diverse in terms of tokens, which can be achieved by paraphrasing the question or introducing/renaming variables"} to the prompts, so the abstract contexts could be generalized. We name our generated dataset from ChatGPT for evaluations as {\dataset}, consisting of $16K$ QA pairs. Our prompt template is below.

\begin{tcolorbox}[
    title=\textcolor{white}{Base prompt for generating pre-university math questions.}]
    Define Difficulty Level is:...\\
    Define the [\textcolor{blue}{subtopic name}] topic as: [\textcolor{blue}{subtopic definition}]. Generate a math problem with its answer at Difficulty Level [\textcolor{blue}{difficulty}] in the topic of [\textcolor{blue}{topic}]: [\textcolor{blue}{subtopic name}]. \\
    Your questions are required to be diverse ...\\
    You are also given an example:
    [\textcolor{blue}{prompt demonstration}]\\
    Generated question:...
\label{base-prompt-format}
\end{tcolorbox}

We also conduct an in-depth analysis of {\dataset} to better understand how large and diverse in terms of topic, lesson, and difficulty our evaluations are. Its statistics are presented in \Cref{tab:diffdist}. Our analysis offers several key insights. First, regarding difficulty, {\dataset} encompasses question-answer pairs from five distinct difficulty levels, with Level 4 being the most prevalent at 54.7\% and Level 3 at 13.5\%. In terms of topic and lesson distribution, grade 1 exhibits the fewest topics (2), while tertiary classes have the highest (51), likely due to the broad subject range. Additionally, lesson distribution mirrors topic distribution, with grade 1 having the least number of lessons and tertiary classes having the most. Regarding the generated QA pairs distribution, we observe that in certain lessons such as \emph{Polynomial factorization} (437 QA pairs) and \emph{Trigonometry} (516 QA pairs), ChatGPT can generate substantial numbers of QA pairs, whilst other lessons such as \emph{Absolute value \& piecewise functions} (19 QA pairs) these numbers are significantly fewer. This is because, in certain lessons, problems can have multiple conditions and mathematical scenarios which result in a high number of variants being generated, while questions in other lessons can be either too narrow or too specific, leading to limited variants. Therefore, the number of generated QA pairs is not always monotonically increasing with the number of lessons. According to grade, we obtain the number of generated QA pairs for tertiary classes as highest (11,032) while for secondary and elementary classes, grades 6 and 8 have the highest numbers whilst grade 1 is the lowest.

\section{Experimentation}
\label{sec:experimentation}

\subsection{Context-aware Experimentation}

\begin{table*}[t!]
\centering
\small 
\scalebox{1}{
    \begin{tabular}{lc|cccc|cccc|cccc}
        \toprule
        \multirow{2}{*}{\textbf{Model}} & \multirow{2}{*}{\textbf{Mode}} & \multicolumn{4}{c}{\textbf{SVAMP}}  & \multicolumn{4}{c}{\textbf{GSM8K}} & \multicolumn{4}{c}{\textbf{MATH}}\\
         &  & B-4 & R-L & Meteor & BERTScore & B-4 & R-L & Meteor & BERTScore & B-4 & R-L & Meteor & BERTScore \\
        \midrule
        GPT-2 &  fine-tune & 7.78 & 39.39 & 38.49  & 82.01 & 16.70 & 49.25 & 49.17 & 92.36 & 15.53 & 35.68 & 37.03 & 85.93 \\
        BART &  fine-tune & 45.37 & 73.78 & 76.70  & 96.34 & 29.93 & 61.28 & 59.49 & 94.50 & \textbf{32.94} & \textbf{57.76} & \textbf{53.69} & \textbf{92.19}\\
        T5 &  fine-tune & 50.90 & 76.44 & \textbf{78.21}  & 96.54 & 33.54 & 63.33 & 64.35 & \textbf{94.85} & 17.56 & 43.66 & 38.71 & 88.86\\
        Mix-QG &  fine-tune & 46.52 & 75.72 & 75.72  & 96.16 & 27.76 & 58.50 & 57.09 & 94.04 & 18.11 & 44.73 & 39.76 & 89.00\\
        Flan-T5 &  fine-tune & \textbf{51.82} & \textbf{77.01} & 77.84  & \textbf{96.56} & 29.34 & 60.12 & 58.51 & 94.22 & 18.03 & 44.32 & 39.46 & 89.03\\
        ProphetNet &  fine-tune & 25.73 & 70.01 & 68.07  & 94.93 & \textbf{48.57} & \textbf{71.91} & \textbf{70.88} & 91.79 & 3.40 & 17.30 & 8.41 & 61.56\\
        \midrule
        GPT-3 &  zero-shot & 1.56 & 17.62 & 21.95  & 89.20 & 7.07 & 32.05 & 31.08 & 90.73 & 14.48 & 39.70 & 36.08 & 88.41\\
        GPT-3.5 &  zero-shot & 25.14 & 64.49 & 64.63  & 94.89 & 18.98 & 48.11 & 50.30 & 93.25 & 23.68 & 45.15 & 49.99 & 90.23\\
        ChatGPT &  zero-shot & 23.57 & 57.82 & 65.43 & 93.97 & 12.19 & 38.55 & 44.27 & 91.49 & 19.48 & 42.12 & 46.36 & 89.11\\
        \bottomrule
    \end{tabular}
}
\caption{
    \small{Answer-aware question generation experimental results in the context-aware setting.}
}
\label{tab:answer-aware-results}
\vspace{-4mm}
\end{table*}

\paragraph{$\bullet$ Datasets}  We use SVAMP \cite{svamp}, GSM8K \cite{cobbe2021training}, and MATH \cite{hendrycks2021measuring} as our math question generation benchmarks. SVAMP suits elementary-level math, GSM8K targets secondary school students whilst MATH encompasses tertiary and olympiad levels, covering a wide range of mathematical topics.



\paragraph{$\bullet$ Data Pre-processing} While the SVAMP and GSM8K datasets provide context and question separately, the MATH dataset lacks this separation. To address this, we firstly segment MATH contexts into individual sentences, then the annotators identify the most suitable sentence for forming a question and the rest becomes context. In cases where the information is insufficient, the annotators can exclude these samples. Finally, in contrast to GSM8K and MATH, which provide separate train and test sets, we split the SVAMP dataset into train and test sets due to the absence of this division.

\paragraph{$\bullet$ Baselines} We use GPT-2 \cite{radford2019language}, BART \cite{lewis-etal-2020-bart}, T5 \cite{2020t5}, MixQG \cite{murakhovska-etal-2022-mixqg}, Flan-T5 \cite{chung2022scaling}, and ProphetNet \cite{qi2020prophetnet} as our fine-tuned question generation baselines. They were initialized with \cite{wolf-etal-2020-transformers} checkpoints and fine-tuned for $10k$ iterations using the AdamW optimizer \cite{loshchilov2018decoupled}. Learning rates of $1e-5$, $5e-5$, and $5e-5$ were respectively used for SVAMP, GSM8K, and MATH.

\paragraph{$\bullet$ Automatic Evaluation} In the \emph{answer-aware} setup, our aim is to generate questions that closely resemble ground-truth one as possible. Following previous works \cite{du-etal-2017-learning, murakhovska-etal-2022-mixqg, do-etal-2022-cohs}, we use BLEU-4 \cite{papineni-etal-2002-bleu}, ROUGE-L \cite{lin-2004-rouge}, METEOR \cite{banerjee-lavie-2005-meteor} as our n-gram evaluation metrics, and use BERTScore \cite{bert-score} to measure the similarity between the generated candidate and ground-truth questions. In the \emph{answer-unaware} setting where the answer and the ground-truth question are unavailable, we follow \cite{shen-etal-2021-gtm, do2023modeling} and measure the Diversity of generated questions by Distinct-1,2 \cite{li-etal-2016-diversity} and the Relevancy with respect to the context using BERTScore.

\paragraph{$\bullet$ Human Evaluation} To further assess the quality of the generated questions with human preferences, we conduct a human study on 200 randomly selected cases from each dataset. The best-performing fine-tuned baseline (based on the average of all metrics) and ChatGPT are selected for evaluation. Then, three English-native educators are hired to evaluate models (1-5) based on 5 criteria: Difficulty, Relevancy, Grammaticality, Answerability, and Usefulness.  The detailed scoring criteria for metrics are provided in \Cref{appdx:human-rating}.



\subsection{Context-unaware Experimentation} \label{sec:context-unaware-experimentation}

In the context-unaware setting, since there are no ground-truth questions, we only rely on human evaluations. We perform human evaluation on 500 randomly selected samples, with 100 questions coming from each {\it prompted} difficulty. We hire three educators who are native English speakers to evaluate ChatGPT on 5 criteria: \textbf{(1) Grammaticality} to assess the grammatical accuracy of the generated question; \textbf{(2) Answerability} measuring the answerable plausibility of the generated question; \textbf{(3) Topic Alignment} assessing question relevancy to the topic; \textbf{(4) Difficulty Alignment} to compare the expected and generated difficulty; \textbf{(5) Usefulness} to assess the mathematical usefulness of the generated question to the education generally. The scoring criteria for metrics are provided in \Cref{appdx:human-rating}.

\subsection{Human Rating System}
\label{appdx:human-rating}
This section presents the human evaluation criteria employed to assess the quality of datasets in both context-aware and context-unaware settings. These criteria were thoughtfully selected, taking into consideration their widespread usage, to ensure an effective evaluation of the datasets' quality.

For evaluating both answer-aware and answer-unaware settings, our human evaluators assess questions based on multiple criteria. These criteria encompass Difficulty, Relevancy, Grammaticality, and Answerability. When evaluating difficulty, we employ a 1 to 5 scale, with 1 signifying suitability for lower elementary school students (grades 1-3) and 5 representing a level of challenge appropriate for mathematics contests and tertiary-level students. Relevancy scores span from 1 to 5, with 1 indicating low relevance (0-20\%) and 5 denoting high relevance (80-100\%) between the context and the generated question. Grammaticality is rated with options of 1, 3, or 5, where 1 reflects the presence of severe grammatical errors, 3 suggests the question is good but contains minor grammatical errors, and 5 indicates a question that is both grammatically and factually correct. As for answerability, we consider two scenarios. In the answer-aware setting, a score of 1 means the question is not answerable, and a score of 3 indicates that the question is answerable but does not match the ground-truth answer, while a score of 5 means the answer matches the ground-truth. In the answer-unaware setting, only a score of 5 is used, indicating that the question is answerable.

To evaluate questions within the PRE-UMATH framework, human evaluators employ a set of diverse criteria, encompassing Difficulty, Grammaticality, Answerability, Topic Alignment, Difficulty Alignment, Answer Quality, and Usefulness. Difficulty is rated from 1 to 5 scale, indicating the question's suitability for varying educational levels, from elementary to olympiad. Grammaticality is scored at 1, 3, or 5, reflecting the presence of grammatical errors. Answerability is rated either 1 (unanswerable) or 5 (answerable). Topic alignment is rated as either 1 (not relevant to either topic or lesson), 3 (relevant to the topic but not the lesson), or 5 (relevant to both topic and lesson). Difficulty alignment is assessed with options of 1 (deviation of more than 1 level from the given difficulty), 3 (one-level deviation), and 5 (match the given difficulty level). Answer Quality is either 1 (incorrect step-by-step explanation), 3 (partially correct step-by-step explanation but not the final answer), or 5 (both a correct step-by-step explanation and the correct final answer). Finally, usefulness scores gauge the utility of generated questions and solutions, which is either 1 (not useful), 3 (useful but requires editing), or 5 (useful and no editing required).

\begin{table}[t!]
\centering
\setlength{\tabcolsep}{4pt}
\small
\scalebox{0.7}{
\begin{tabular}{lc|ccc|ccc|ccc}
\toprule
\multirow{2}{*}{\textbf{Model}} & \multirow{2}{*}{\textbf{Mode}} & \multicolumn{3}{c}{\textbf{SVAMP}}  & \multicolumn{3}{c}{\textbf{GSM8K}} & \multicolumn{3}{c}{\textbf{MATH}}\\
 &  & Dist-1 & Dist-2 & Rel. & Dist-1 & Dist-2 & Rel. & Dist-1 & Dist-2 & Rel. \\
\midrule
GPT-2 &  fine-tune & \textbf{21.88} & \textbf{61.89} & 86.28  & 12.72 & 50.76 & 86.53 & 4.72 & 19.40 & 81.08 \\
BART &  fine-tune & 16.07 & 41.34 & 88.11 & 15.45 & 46.64 & 87.13 & 4.62 & 16.20 & \textbf{88.80} \\
T5 &  fine-tune &  15.80 & 41.85 & 88.22  & 15.67 & 46.96 & 87.54 & 4.05 & 12.68 & 88.48 \\
Mix-QG &  fine-tune & 15.33 & 39.14 & 88.30  & 16.59 & 47.48 & 87.56 & 4.84 & 14.92 & 83.33\\
Flan-T5 &  fine-tune & 15.63 & 40.15 & 88.28  & 16.17 & 46.97 & 87.56 & 3.35 & 10.39 & 83.03 \\
ProphetNet &  fine-tune & 16.35 & 39.60 & 88.01  & 9.28 & 35.76 & 85.54 & 2.29 & 16.36 & 59.62\\

\midrule
GPT-3 &  zero-shot & 17.36 & 47.10 & 88.81  & 15.65 & 51.79 & 87.85 & 10.62 & 31.50 & 84.85 \\
GPT-3.5 &  zero-shot & 17.48 & 43.02 & 87.76  & 16.31 & 50.55 & 87.36 & 10.32 & 29.37 & 83.94 \\
ChatGPT &  zero-shot & 19.91 & 51.67 & \textbf{89.40}  & \textbf{17.15} & \textbf{56.11} & \textbf{88.28} & \textbf{11.27} & \textbf{35.66} & 86.25\\
\bottomrule
\end{tabular}
}
\caption{\small{Answer-unaware question generation results in the context-aware setting.}}
\label{tab:answer-unaware-results}
\vspace{-10mm}
\end{table}

\begin{table*}[t!]
\centering
\small 
\scalebox{.8}{
\begin{tabular}{l|cccccc|cccccc|cccccc}
\toprule
\multirow{2}{*}{\textbf{Setup}} & \multicolumn{6}{c}{\textbf{SVAMP}}  & \multicolumn{6}{c}{\textbf{GSM8K}} & \multicolumn{6}{c}{\textbf{MATH}}\\
 & Model & Diff. & Rel. & Gram. & Ans. & Use. & Model & Diff. & Rel. & Gram. & Ans. & Use. &  Model & Diff. & Rel. & Gram. & Ans. & Use. \\
\midrule
Answer-aware & Flan-T5 & 1.03 & \textbf{4.99} & 4.91 & 4.01 & 3.65 & T5 & 1.91 & 4.97 & 4.89 & 4.27 & 3.89 & BART & 3.56 & 4.82 & 4.91 & 3.79 & 3.59 \\
Answer-aware & GPT-3.5 & 1.04 & \textbf{4.99}  & 4.96 & 4.17 & 4.44 & GPT-3.5 & \textbf{2.04} & \textbf{4.98} & 4.96 & 3.51 & 3.95 & GPT-3.5 & 3.57 & 4.84 & \textbf{4.97} & 4.07 & 3.92 \\
Answer-aware & ChatGPT & 1.06 & \textbf{4.99} & 4.96 & 4.39 & \textbf{4.73} & ChatGPT & 1.96 & \textbf{4.98} & \textbf{4.98} & 3.94 & \textbf{4.31} & ChatGPT & \textbf{3.73} & 4.90 & 4.94 & \textbf{4.53} & \textbf{4.19}\\
\midrule
Kripp's Alpha &  & 61.01 & 69.71  & 70.23 & 76.37 & 65.46 & & 70.98 & 75.12 & 78.53 & 83.68 & 67.91 & & 51.25 & 58.83 & 66.52 & 63.09 & 60.09 \\
\midrule
\midrule
Answer-unaware & GPT-2 & 1.02 & 4.53 & 4.06 & 3.41 & 2.96 & Mix-QG & 1.66 & 4.92 & \textbf{4.98} & 3.42 & 3.47 & GPT2 & 1.02 & 3.21 & 2.46 & 1.53 & 2.09 \\
Answer-unaware & GPT-3 & \textbf{1.14} & 4.82  & 4.96 & \textbf{4.84} & 4.27 & GPT-3 & 1.66 & 4.95 & 4.85 & 3.81 & 3.61 & GPT-3 & 3.44 & 4.81 & 4.85 & 4.16 & 3.75 \\
Answer-unaware & ChatGPT & 1.06 & \textbf{4.99}  & \textbf{4.98} & 4.81 & 4.61 & ChatGPT & 1.74 & 4.96 & \textbf{4.98} & \textbf{4.69} & 4.14 & ChatGPT & 3.47 & \textbf{4.98} & 4.92 & 4.24 & 4.07 \\
\midrule
Kripp's Alpha &  & 66.23 & 71.18  & 75.56 & 74.09 & 58.43 &  & 64.33 & 71.57 & 73.68 & 75.09 & 60.55 &  & 64.26 & 70.01 & 71.23 & 65.09 & 69.91\\
\bottomrule
\end{tabular}
}
\caption{\small{Human evaluation results in the context-aware setting.}}
\label{tab:human-evaluation-context-aware}
\vspace{-3mm}
\end{table*}

\begin{table}[ht]
\centering
\small
\scalebox{0.7}{
    \begin{tabular}{p{1.5cm}|p{1.5cm}p{1.5cm}p{1.5cm}p{1.5cm}p{1.5cm}}
    \toprule
        Class & Q. Gram. & Q. Ans. & Top. Ali. & Dif. Ali. & Use. \\
        \midrule
        Grade 1 & \textbf{5.00 $\pm$ 0.00} & 4.60 $\pm$ 1.26 & 4.60 $\pm$ 1.26 & \textbf{5.00 $\pm$ 0.00} & 3.80 $\pm$ 1.69 \\ 
        Grade 2 & \textbf{5.00 $\pm$ 0.00} & \textbf{4.92 $\pm$ 0.55} & \textbf{4.92 $\pm$ 0.55} & \textbf{5.00 $\pm$ 0.00} & 4.55 $\pm$ 1.15\\ 
        Grade 3 & 4.82 $\pm$ 0.58 & 4.84 $\pm$ 0.78 & 4.84 $\pm$ 0.63 & 4.95 $\pm$ 0.32 & 4.35 $\pm$ 1.32\\ 
        Grade 4 & 4.92 $\pm$ 0.40 & \textbf{4.92 $\pm$ 0.57} & 4.84 $\pm$ 0.69 & 4.59 $\pm$ 0.81 & 4.22 $\pm$ 1.46\\ 
        Grade 5 & 4.86 $\pm$ 0.60 & 4.81 $\pm$ 0.86 & 4.83 $\pm$ 0.56 & 4.76 $\pm$ 0.65 & 4.43 $\pm$ 1.23\\ 
        Grade 6 & 4.90 $\pm$ 0.44 & 4.73 $\pm$ 1.01 & 4.80 $\pm$ 0.80 & 4.14 $\pm$ 1.27 & \textbf{4.56 $\pm$ 1.12}\\ 
        Grade 7 & 4.88 $\pm$ 0.48 & 4.88 $\pm$ 0.69 & 4.71 $\pm$ 1.00 & 4.32 $\pm$ 1.22 & 4.18 $\pm$ 1.31\\ 
        Grade 8 & 4.67 $\pm$ 0.85 & 4.18 $\pm$ 1.63 & 4.71 $\pm$ 0.91 & 4.67 $\pm$ 0.75 & 4.02 $\pm$ 1.42\\ 
        Tertiary & 4.88 $\pm$ 0.53 & 4.55 $\pm$ 1.26 & 4.89 $\pm$ 0.60 & 4.50 $\pm$ 0.99 & 4.29 $\pm$ 1.24\\
        \midrule
        \dataset & 4.87 $\pm$ 0.54 & 4.68 $\pm$ 1.09 & 4.84 $\pm$ 0.68 & 4.61 $\pm$ 0.90 & 4.32 $\pm$ 1.27 \\
        \bottomrule
    \end{tabular}
}
\caption{Human evaluation results in the context-unaware setting. The format is $\mu + \sigma$}
\label{tab:my_label}
\vspace{-7mm}
\end{table}

\section{Results and Discussions}
\label{sec:result-discussion}
\subsection{Automatic Evaluation}

It is worth noting that our automatic evaluations are only conducted on \emph{context-aware} setting. In the \emph{answer-aware} setting, fine-tuning baselines consistently outperform ChatGPT across all automatic metrics on the three benchmarks. However, in the \emph{answer-unaware} setting, we derive interesting insights. Firstly, ChatGPT generates more diverse questions in terms of token levels on the challenging GSM8K and MATH benchmarks, which underscores its practical potential for educational purposes. Conversely, GPT-2 excels on SVAMP dataset by yielding higher distinct scores compared to ChatGPT. This might be because the questions generated by GPT-2 are generally short and consist of non-sense tokens. For example: \texttt{Context: ``At the stop 8 more people got on the train. There were 11 people on the train.''; Question: ``@@ Is there a limit on the number of people on the bus?''}



\subsection{Context-aware Human Evaluation} \label{subsec:context-aware-human-eval}



Through our careful manual evaluations, we have obtained 6 insightful findings.

\paragraph{(1) ChatGPT generates questions with minimal grammatical errors.} 
As shown in \Cref{tab:human-evaluation-context-aware}, ChatGPT consistently attains grammaticality scores exceeding $>4.9$, underscoring its proficiency in generating grammatically correct texts across all pre-university levels. Notably, we observe that grammatical errors predominantly emerge when ChatGPT attempts to generate highly complex problems.

\paragraph{(2) ChatGPT generates questions that are highly relevant to the input context.} Our manual evaluations reveal that the questions generated by ChatGPT are highly relevant to the input contexts. Remarkably, these questions exhibit minimal presence of unrelated characters or variables not found in the context, resulting in nearly perfect relevancy scores across most datasets and sub-settings (see \Cref{tab:human-evaluation-context-aware}). However, an intriguing observation emerges with a lower relevancy score in the answer-aware setting for MATH compared to its answer-unaware counterpart.

\paragraph{(3) ChatGPT frequently repeats information from the context} 
Despite explicit constraints outlined in the prompt template regarding repetition, the model occasionally reiterates random segments (tertiary-level) or the whole context (lower-levels). This repetition leads to the generation of overly lengthy questions, as exemplified by instances such as: ``Context: \emph{A football team played 22 games. They won 8 more than they lost.''}, ``Generated Question: \emph{How many games did the football team win if they played 22 games and won 8 more than they lost?''}. Subsequent human evaluations unveil that this issue predominantly afflicts the GSM8K dataset, occurring about 50\% of the time.

\paragraph{(4) With an expected answer, ChatGPT tends to generate answerable questions whilst, without it, this likelihood is lower.} When additional hypotheses are required to construct a complete question (e.g., ``Context: \emph{Mary is two years younger than Joan, who is five years older than Jessa''}), our empirical evaluation indicates that ChatGPT tends to struggle in the absence of an expected answer as a reference. For instance, within the answer-unaware setup and considering the context mentioned, ChatGPT only asks \emph{``How old is Jessa?''}.

\paragraph{(5) Without an expected answer, ChatGPT frequently generates trivial questions} In the \textit{answer-unaware} scenario, ChatGPT often exhibits a combination of the aforementioned behaviors (2) and (4), where it redundantly repeats information from the context and formulates it as a question. This behavior occurs irrespective of the context's complexity, resulting in the generation of simplistic questions that merely require looking up information in the context itself. For instance, when the context is as straightforward as \emph{``Darrell and Allen's ages are in the ratio of 7:11''}, ChatGPT redundantly repeats the entire context and asks, \emph{``What is the ratio of Darrell’s age to Allen’s age?''}. While this phenomenon occurs less frequently than (4), about 5\% of the generated questions.

\paragraph{(6) Even with good contextual understanding, ChatGPT struggles to understand the relationship between mathematical objects.} This problem manifests in both the \emph{answer-aware} and \emph{answer-unaware} settings. In the \emph{answer-aware} mode, ChatGPT tends to inaccurately order subtraction operations while in the \emph{answer-unaware}, it exhibits reluctance in generating questions related to the relationships between objects. This phenomenon occurs about 2\% of the time.


\subsection{Context-unaware Human Evaluation} \label{subsec:context-unaware-human-eval}

Along with the same observation about question grammaticality, we provide 5 more findings in the {\it context-unaware} setting.

\paragraph{(1) ChatGPT generates questions with high diversity in terms of context.} Across all three class levels, ChatGPT consistently excels at crafting real-life scenarios that seamlessly integrate with the context. Notably, at the secondary and tertiary school levels, ChatGPT demonstrates its versatility by drawing connections to other subjects, such as Physics and Biology. For example: \emph{``A population of bacteria doubles every 3 hours. If there are initially 1000 bacteria, what will be the population after 12 hours? Is this an example of exponential growth or decay?''}

\paragraph{(2) ChatGPT sometimes generates questions that are not well-aligned with some provided topics.} ChatGPT occasionally interprets complex concepts as more familiar ones. For instance, when tasked with creating questions related to \emph{``Divide whole numbers to get a decimal quotient''}, it might generate questions whose answers are whole numbers instead of decimals. Similarly, in lessons containing the term \emph{``modeling''}, ChatGPT tends to generate Linear Programming questions. Although this misunderstanding is relatively infrequent, only about 5\% of cases, it is a noteworthy aspect to keep in mind.

\paragraph{(3) ChatGPT generates questions with difficulty solely depending on the difficulty of demonstration.} When presented with a straightforward example (level 3) but tasked with generating a more complex question (level 5), ChatGPT often replicates the initial demonstration and struggles to enhance the question's difficulty. We noticed that lessons with overlapping content between secondary school and high school levels were generated similarly because of the shared demonstration, despite distinct required difficulty levels in the prompt. This pattern was consistently observed in all our attempts to generate questions with varying difficulty levels from the provided demonstrations.

\paragraph{(4) If ChatGPT generates hard questions, it could not handle the complexity and generates nonsense.} ChatGPT demonstrates proficiency in introducing new objects within questions but struggles to establish meaningful connections or inquire about genuine relationships between these objects. For instance, in Geometry questions, ChatGPT often generates random points (A, B, C), states some relationships (e.g., AX is the bisector of angle BAC), and includes unrelated quantitative properties (e.g., angle AXB = angle AXC), resulting in suboptimal performance.

\paragraph{(5) ChatGPT could generate questions that are not so mathy.} While instructions and demonstrations in prompts have been effective in mitigating the issue, they are not foolproof. In primary school topics such as \emph{``Measurement''}, ChatGPT occasionally generates questions that do not necessitate mathematical knowledge (e.g., \emph{``How long is a ruler?''}). However, in higher-level classes where lesson names are more math-specific, this phenomenon is notably less prevalent.

\section{Conclusions}
In this work, we provide an in-depth analysis of the capability of ChatGPT to generate pre-university math questions. Our experimentation is categorized into two main settings: \emph{(1) context-aware} - when an input context/background is provided; \emph{(2) context-unaware} - when there is no available context provided. The evaluation results in the context-aware setting show that although the generated questions are highly grammatically correct and relevant to the context, they tend to have lower difficulty than expected. Additionally, we find that with an expected answer as input, ChatGPT is more likely to generate an answerable question than without any answer provided. In the context-unaware setting, to exhaustively evaluate ChatGPT in pre-university math topics, we first crawl \topicmath{}, an expert-authored pre-university math curriculum consisting of $121$ math topics and $428$ lessons with their definitions and question-answer examples. We then prompt ChatGPT to generate math questions within each topic. Our human evaluations reveal that in some topics, the generated questions are not well-aligned with the topics, and are combined with knowledge from other scientific fields. Furthermore, we find that by providing both difficulty requirements and demonstrations, ChatGPT is highly likely to generate questions that are aligned with demonstrations instead of difficulty instructions. We hope these findings can provide good insights for teachers and researchers in utilizing large language models such as ChatGPT for generating math questions, boosting the applications of AI in education.

\section*{Acknowledgements}
This research/project is supported by the National Research Foundation, Singapore under its AI Singapore Programme, AISG Award No: AISG2-TC-2022-005, and by MoE Tier 1 research grant number RS21/20, Singapore. Do Xuan Long is supported by the A*STAR Computing and Information (ACIS) scholarship.

\appendix

\bibliographystyle{ACM-Reference-Format}
\bibliography{sample-bibliography} 

\end{document}